\documentclass[runningheads]{llncs}
\usepackage{graphicx}
\usepackage{amsfonts}
\usepackage{float}
\usepackage{multirow}
\begin{document}
\title{A study of text representations for Hate Speech Detection
  \thanks{Supported by NCSR Demokritos, and the Department of Informatics and Telecommunications, National and Kapodistrian University of Athens}
}
%
%

\author{Chrysoula Themeli\inst{1} \and
George Giannakopoulos\inst{2} \and
Nikiforos Pittaras\inst{1,2}}

\authorrunning{Ch. Themeli et al.}



%
\institute{Department of Informatics and Telecommunications, \\ National and Kapodistrian University of Athens \\
\email{\{cthemeli, npittaras\}@di.uoa.gr} \and
NCSR Demokritos, Athens, Greece \\
\email{\{ggianna, pittarasnikif\}@iit.demokritos.gr}
}

 
%
\maketitle              
\begin{abstract}
The pervasiveness of the Internet and social media have enabled the rapid and anonymous spread of Hate Speech content on microblogging platforms such as Twitter. Current EU and US legislation against hateful language, in conjunction with the large amount of data produced in these platforms has led to automatic tools being a necessary component of the Hate Speech detection task and pipeline. In this study, we examine the performance of several, diverse text representation techniques paired with multiple classification algorithms, on the automatic Hate Speech detection and abusive language discrimination task. 
We perform an experimental evaluation on binary and multiclass datasets, paired with significance testing. Our results show that simple hate-keyword frequency features (BoW) work best, followed by pre-trained word embeddings (GLoVe) as well as N-gram graphs (NGGs): a graph-based representation which proved to produce efficient, very low-dimensional but rich features for this task. A combination of these representations paired with Logistic Regression or 3-layer neural network classifiers achieved the best detection performance, in terms of micro and macro F-measure. 
\keywords{hate speech \and natural language processing \and classification \and social media}
\end{abstract}

\section{Introduction}

Hate Speech is a common affliction in modern society. Nowadays, people can come across Hate Speech content even more easily through social media platforms, websites and forums containing user-created content. The increase of the use of social media gives individuals the opportunity to easily spread hateful content and reach a number of people larger than ever before. On the other hand, social media platforms like Facebook or Twitter want to both comply with legislation against Hate Speech and improve user experience. Therefore, they need to track and remove Hate Speech content from their websites efficiently.

Due to the large amount of data transmitted through these platforms, delegating such a task to humans is extremely inefficient. A usual compromise is to rely on user reports in order to review only the reported posts and comments. This is also ineffective, since it relies on the users' subjectivity and trustworthiness, as well as  their ability to thoroughly track and flag such content. Due to all of the above, the development of automated tools to detect Hate Speech content is deemed necessary.
The goal of this work is: (i) to study different text representations and classification algorithms in the task of Hate Speech detection; (ii) evaluate whether the n-gram graphs representation \cite{giannakopoulos2008testing} can constitute a rich/deep feature set (as e.g. in \cite{papadakis2016graph}) for the given task. 

The structure of the paper is as follows. In section \ref{sec:ProblemDefinition} we define the hate speech detection problem, while in section \ref{sec:RelatedWork} we discuss related work. We overview our study approach and elaborate on the proposed method in section \ref{sec:ProposedMethod}. We then experimentally evaluate the performance of different approaches in Section \ref{sec:Experiments}, concluding the paper in Section \ref{sec:Conclusion}, by summarizing the findings and proposing future work.

\section{Problem Definition}\label{sec:ProblemDefinition}


The first step to Hate Speech detection is to provide a clear and concise definition of Hate Speech. This is important especially during the manual compilation of Hate Speech detection datasets, where human annotators are involved. In their work, the authors of \cite{kwok2013locate} have asked three students of different race and same age and gender to annotate whether a tweet contained Hate Speech or not, as well as the degree of its offensiveness. The agreement was only $33\%$, showing that Hate Speech detection can be highly subjective and dependent on the educational and/or cultural background of the annotator. Thus, an unambiguous definition is necessary to eliminate any such personal bias in the annotation process.

Usually, Hate Speech is associated with insults or threats. Following the definition provided by \cite{brown2017hate}, ``it covers all forms of expressions that spread, incite, promote or justify racial hatred, xenophobia, antisemitism or other forms of hatred based on intolerance''. Moreover, it can be ``insulting, degrading, defaming, negatively stereotyping or inciting hatred, discrimination or violence against people in virtue of their race, ethnicity, nationality, religion, sexual orientation, disability, gender identity''. However, we cannot disregard that Hate Speech can be also expressed by statements promoting superiority of one group of people against another, or by expressing stereotypes against a group of people.

The goal of a Hate Speech Detection model is, given an input text $T$, to output \texttt{True}, if $T$ contains Hate Speech and \texttt{False} otherwise. Modeling the task as a binary classification problem, the detector is built by learning from a training set and is subsequently evaluated on unseen data. Specifically, the input is transformed to a machine-readable format via a text representation method, which ideally captures and retains informative characteristics in the input text. The representation data is fed to a machine learning algorithm that assigns the input to one of the two classes, with a certain confidence. During the training phase, this discrimination information is used to construct the classifier. The classifier is then applied on data not encountered during training, in order to measure its generalization ability.

In this study, we focus on user-generated texts from social media platforms, specifically Twitter posts. We evaluate the performance of several established text representations (e.g. Bag of words, word embeddings) and classification algorithms. We also investigate the contribution of the graph-based the n-gram graph features to the Hate Speech classification process. Moreover, we examine whether a combination of deep features (such as n-gram graphs) and shallow features (such as Bag of Words) can provide top performance in the Hate Speech detection task.

\section{Related Work}\label{sec:RelatedWork}

In this section, we provide a short review of the related work, not only for Hate Speech detection, but for similar tasks as well. Examples of such tasks can be found in \cite{mubarak2017abusive} where the authors aim to identify which users express Hate Speech more often, while \cite{xu2010filtering} detect and delete hateful content in a comment, making sure what is left has correct syntax. The latter is a demanding task which requires the precise identification of grammatical relations and typed dependencies among words of a sentence. Their proposed method results have $90.94\%$ agreement with the manual filtering results. 

Automatic Hate Speech detection is usually modeled as a Binary Classification. However, multi-class classification can be applied to identify the specific kind of Hate Speech (e.g. racism, sexism etc) \cite{badjatiya2017deep,park2017one}. One other useful task is the detection of the specific words or phrases that are offensive or promote hatred, investigated in \cite{warner2012detecting}. 

\subsection{Text representations for Hate Speech}
In this work we focus on representations, i.e. the mapping of written human language into a collection of useful features in a form that is understandable by a computer and, by extension, a Hate Speech Detection model. Below we overview a number of different representations used within this domain. 

A very popular representation approach is the Bag of Words (BOW) \cite{kwok2013locate,bourgonje2017automatic,badjatiya2017deep} model, a Vector Space Model extensively used in Natural Language Processing and document classification. In BOW, the text is segmented to words, followed by the construction of a histogram of (possible weighted) word frequencies. Since BOW discards word order, syntactic, semantic and grammatical information, it is commonly used as a baseline in NLP tasks.
An extension of the BOW is the Bag of N-grams \cite{nobata2016abusive,kwok2013locate,mubarak2017abusive,davidson2017automated,waseem2016you}, which replaces the unit of interest in BOW from words to $n$ contiguous tokens. A token is usually a word or a character in the text, giving rise to word n-gram and character n-gram models. Due to the contiguity consideration, n-gram bags retain local spacial and order information.

The authors in \cite{davidson2017automated} claim that lexicon detection methods alone are inadequate in distinguishing between Hate Speech and Offensive Language, counter-proposing n-gram bags with TF-IDF weighting along with a sentiment lexicon, classified with L2 regularized Logistic Regression \cite{mccullagh1989generalized}. On the other hand, \cite{badjatiya2017deep} use character n-grams, BOW and TF-IDF features as a baseline, proposing word embeddings from GloVe \footnote{\url{https://nlp.stanford.edu/projects/glove/}}.
In \cite{park2017one} the authors use character and word CNNs as well a hybrid CNN model to classify sexist and racist Twitter content. They compare multi-class detection with a coarse-to-fine two-step classification process, achieving similar results with both approaches. There is also a variety of other features used such as word or paragraph embeddings (\cite{djuric2015hate}, \cite{warner2012detecting}, \cite{badjatiya2017deep}), LDA and Brown Clustering (\cite{saleem2017web}, \cite{xiang2012detecting}, \cite{warner2012detecting}, \cite{waseem2016you}), sentiment analysis(\cite{gitari2015lexicon}, \cite{davidson2017automated}), lexicons and dictionaries (\cite{gitari2015lexicon}, \cite{silva2016analyzing}, \cite{del2017hate} etc) and POS tags(\cite{nobata2016abusive}, \cite{xu2010filtering}, \cite{saleem2017web} etc).

\subsection{Classification approaches}
Regarding classification algorithms, SVM \cite{cortes1995support}, Logistic Regression (LR) and Naive Bayes (NB) are the most widely used (e.g. \cite{warner2012detecting,saleem2017web,davidson2017automated,djuric2015hate} etc). In \cite{waseem2016hateful} and \cite{xiang2012detecting}, the authors use a bootstrapping approach to aid the training process via data generation. This approach was used as a semi-supervised learning process to generate additional data automatically or create hatred lexical resources. The authors of \cite{xiang2012detecting} use the Map-Reduce framework in Hadoop to collect tweets automatically from users that are known to use offensive language, and a bootstrapping method to extract topics from tweets.

Other algorithms used are Decision Trees and Random Forests (RF) (\cite{davidson2017automated,bourgonje2017automatic,xiang2012detecting}), while \cite{badjatiya2017deep} and \cite{del2017hate} have used Deep Learning approaches via LSTM networks. Specifically, \cite{badjatiya2017deep} use CNN, LSTM and FastText, i.e. a model that is represented by average word vectors similar to BOW, which are updated through backpropagation. The LSTM model achieved the best performance with $0.93$ F-Measure, used to train a GBDT (Gradient Boosted Decision Trees) classifier. In \cite{davidson2017automated}, the authors use several classification algorithms such as regularized LR, NB, Decision Trees, RF and Linear SVM, with L2-regularized LR outperforming other approaches in terms of F-score.

For more information, the survey of \cite{schmidt2017survey} provides a detailed analysis of detector components used for Hate Speech detection and similar tasks. 

\section{Study and Proposed Method}
\label{sec:ProposedMethod}

In this section we will describe the text representations and classification components used in our implementations of a Hate Speech Detection pipeline. We have used a variety of different text representations, i.e. bag of words, embeddings, n-grams and n-gram graphs and tested these representations with multiple classification algorithms. We have implemented the feature extraction in Java and used both Weka and scikit-learn (sklearn) to implement classification algorithms. For artificial neural networks (ANNs), we have used sklearn and Keras frameworks. Our model can be found in our GitHub repository \footnote{\url{https://github.com/cthem/hate-speech-detection}}.

\subsection{Text representations}

In order to discard noise and useless artifacts we apply standard preprocessing to each tweet. First, we remove all URLs, mentions (e.g. @username), RT (Retweets) and hashtags (e.g. words starting with \#), as well as punctuation, focusing on the text portion of the tweet. Second, we convert tweets to lowercase and remove common English stopwords using a predefined collection \footnote{\url{https://github.com/igorbrigadir/stopwords}}. 

After preprocessing, we apply a variety of representations, starting with the Bag of Words (BOW) model. 
This representation results in a high dimensional vector, containing all encountered words, requiring a significant amount of time in order to process each text. In order to reduce time and space complexity, we limit the number of words of interest to keywords from HateBase \footnote{\url{https://github.com/t-davidson/hate-speech-and-offensive-language}} \cite{davidson2017automated}.

Moreover, we have used additional bag models, with respect to word and character n-grams. In order to guarantee a common bag feature vector dimension across texts, we pre-compute all n-grams that appear in the dataset, resulting in a sparse and high-dimensional vector.
Similarly to the BOW features, in order to reduce time and space complexity, it is necessary to reduce the vector space. Therefore, we keep only the 100 most frequent n-grams features, discarding the rest.
Unfortunately, as we will illustrate in the experiments, this decision resulted in highly sparse vectors and, thus, reduced the efficiency of those features.

Furthermore, we have used GloVe word embeddings \cite{pennington2014glove} to represent the words of each tweet, mapping each word to a 50-dimensional real vector and arriving at a single tweet vector representation via mean averaging. Words missing from the GloVe mapping were discarded.

Expanding the use of n-grams, we examine wether n-gram graphs (NGGs) \cite{giannakopoulos2009automatic,giannakopoulos2012representation} can have a significant contribution in detecting Hate Speech. NGGs are a graph-based text representation method that captures both frequency and local context information from text n-grams (as opposed to frequency-only statistics that bag models aggregate). This enables NGGs to differentiate between morphologically similar but semantically different words, since the information kept is not only the specific n-gram but also its context (neighboring n-grams).
The graph is constructed with n-grams as nodes and local co-occurence information embedded in the edge weights, with comparisons defined via graph-based similarity measures \cite{giannakopoulos2009automatic}. NGGs can operate with word or character n-grams -- in this work we employ the latter version, which has been known to be resilient to social media text noise \cite{papadakis2016graph,giannakopoulos2012representation}.

During training, we construct a representative category graph (RCG) for each category in the problem (e.g. ``Hate Speech'' or ``Clean''), aggregating all training instances per category to a single NGG. We then compare the NGG of each instance to each RCG, extracting a score expressing the degree that the instance belongs that class -- for this, we use the NVS measure \cite{giannakopoulos2009automatic}, which produces a similarity score between the instance and category NGGs. After this process completes, we end up with similarity-based, $n$-dimensional model vector features for each instance -- where $n$ is the number of possible classes.
We note that we use $90\%$ of the training instances to build the RCGs, in order to avoid overfitting of our model: in short, using all training instances would result in very high instance-RCG similarities during training. Since we use the resulting model vectors as inputs to a classification phase in the next step, the above approach would introduce extreme overfit to the classifier, biasing it towards expecting perfect similarity scores in cases of an instance belonging to a class, a scenario which of course rarely -- if ever --  happens with real world data.

In addition, we produce sentiment, syntax and spelling features. Sentiment analysis could be a meaningful feature, since hatred is related with a negative polarity. For sentiment and syntax feature extraction we use the Stanford NLP Parser \footnote{\url{https://nlp.stanford.edu/software/lex-parser.html}}. This tool performs sentiment extraction of the longest phrase tracked in the input and additionally can be used to provide a syntactic score with syntax trees, corresponding the best attained score for the entire tweet.

Finally, a spelling feature was constructed to examine whether Hate Speech is correlated to the user's proficiency in writing. We have used an English dictionary to collect all English words with correct spelling and, then, for each word in a tweet, we have calculated its edit distance from each word in the dictionary, keeping the smallest value (i.e. the distance from the best match). The final feature kept was the average edit distance for the entire post, with its value being close to $0$ for tweets with the majority of words correctly spelled.
At the end of this process, we obtain a 3-dimensional vector, each coordinate corresponding to the sentiment, syntax and spelling scores of the text.

\subsection{Classification Methods}

Generated features are fed to a classifier that decides the presence of Hate Speech content. We use a variety of classification models, as outlined below. 

Naive Bayes (NB) \cite{russell1995modern} is a simple probabilistic classifier, based on Bayesian statistics. NB makes the strong assumption that instance features are independent from one another, but yields performance comparable to far more complicated classifiers -- this is why it commonly serves as baseline for various machine learning tasks \cite{lewis1998naive}. Additionally, the independence assumption simplifies the learning process, reducing it to the model learning the attributes separately, vastly reducing time complexity on large datasets. 

Logistic Regression (LR) \cite{menard1995applied} is another statistical model commonly applied as a baseline in binary classification tasks. It produces a prediction via a linear combination of the input with a set of weights, passed through a logistic function which squeezes scores in the range between 0 and 1, i.e. thus producing binary classification labels. Training the model involves discovering optimal values for the weights, usually acquired through a maximum likelihood estimation optimization process.

The K-Nearest Neighbor (KNN) classifier \cite{fix1952discriminatory} is another popular technique applied to classification. It is a lazy and non-parametric method; no explicit training and generalization is performed prior to a query to the classification system, and no assumption is made pertaining to the probability distribution that the data follows. Inference requires a defined distance measure for comparing two instances, via which closest neighbors are extracted. The labels of these neighbors determine, through voting, the predicted label of a given instance.

The Random Forest (RF) \cite{liaw2002classification} is an ensemble learning technique used for both classification and regression tasks. It combines multiple decision trees during the training phase by bootstrap-aggregated ensemble learning, aiming to alleviate noise and overfitting by incorporating multiple weak learners. Compared to decision trees, RF produce a split when a subset of the best predictors is randomly selected from the ensemble. 

Artificial Neural Networks (ANNs) are computational graphs inspired by the biological nervous systems. They are composed of a large number of highly interconnected neurons, usually organized in layers in a feed-forward directed acyclic graph. Similarly to a LR unit, neurons compute the linear combination of their input (including a bias term) and pass the result through a non-linear activation function. Aggregated into an ANN, each neuron computes a specific feature from its input, as dictated by the values of the weights and bias. 
ANNs are trained with respect to a loss function, which defines an error gradient by which all parameters of the ANN are shifted. With each optimization step, the model moves towards an optimum parameter configuration. The gradient with respect to all network parameters is computed by the back-propagation method.In our case, we have used an ANN composed of 3 hidden layers with dropout regularization.

\section{Experiments and Results}
\label{sec:Experiments}
In this section, we present the experimental setting used to answer the following:
\begin{itemize}
    \item Which features have the best performance?
    \item Does feature combination improve performance?
    \item Do NGGs have significant / comparable performance to BOW or word embeddings despite being represented by low dimensional vectors?
    \item Are there classifiers performing statistically significantly better than others? Is the selection of features or classifiers more significant, when determining the pipeline for Hate Speech detection?
\end{itemize}
In the following paragraphs, we elaborate on the  the datasets utilized, present experimental and statistical significance results, as well as discuss of our findings. 

\subsection{Datasets and Experimental Setup}

We use the datasets provided by \cite{waseem2016hateful} \footnote{\url{https://github.com/ZeerakW/hatespeech}} and \cite{davidson2017automated} \footnote{\url{https://github.com/t-davidson/hate-speech-and-offensive-language}}. We will refer to the first dataset as RS (racism and sexism detection) and to the second as HSOL (distinguish Hate Speech from Offensive Language).
In both works, the authors perform a multi-class classification task against the corpora. In \cite{waseem2016hateful}, their goal is to distinguish different kinds of Hate Speech, i.e. racism and sexism, and therefore the possible classes in RS are \texttt{Racist}, \texttt{Sexist} or \texttt{None}. In \cite{davidson2017automated}, the annotated classes are \texttt{Hate Speech}, \texttt{Offensive Language} or \texttt{Clean}.

Given the multi-class nature of these datasets, we combine them into a single dataset, keeping only instances labeled \texttt{Hate Speech} and \texttt{Clean} in the original.
We use the combined (RS + HSOL) dataset to evaluate our model implementations on the binary classification task. Furthermore, we run multi-class experiments on the original datasets for completeness, the results of which are omitted due to space limitations, but are available upon request.

We perform three stages of experiments. First, we run a preliminary evaluation on each feature separately, to assess its performance. 
Secondly, we evaluate the performance of concatenated feature vectors, in three different combinations: 1) the top individually performing features by a significant margin (\texttt{best}), 2) all features \texttt{all} and 3) vector-based features (\texttt{vector}), i.e. excluding NGGs. Via the latter two scenarios, we investigate whether NGGs can achieve comparable performance to vector-based features of much higher dimensionality. 

Given the imbalanced dataset used (24463 Hate Speech and 14548 clean samples), we report performance in both macro and micro F-measure.
Finally, we evaluate (with statistical significance testing) the performance difference between run components, through a series of ANOVA and Tukey HSD test evaluations.

\subsection{Results}

Here we provide the main experimental results of our described in the previous section, presented in micro/macro F-measure scores. More detailed results, including multi-class classification are omitted due to space limitations but are available upon request.

Firstly, to answer the question on the value of different feature types, we perform individual runs which designate BOW, glove embeddings and NGG as the top performers, with the remaining features (namely sentiment, spelling / syntax analysis and n-grams) performing significantly worse. All approaches however surpass a baseline performance in terms of a naive majority-class classifier (scoring $0.382 / 0.473$, in terms of macro and micro F-measure respectively) and are described below.
Sentiment, spelling and syntax features proved to be insufficient information sources to the Hate Speech detection classifiers when used separately -- not surprisingly, since they produce one-dimensional features. The best performers are syntax with NNs in terms of micro F-measure ($0.633$) and spelling with NNs in terms of macro F-measure ($0.566$). In contrast n-gram graph similarity-based features perform close to the best performing BOW configuration (cf. Table \ref{table:results}), having just one additional dimension. This implies that appropriate, deep / rich features can still offer significant information, despite the low dimensionality. NGG-based features appear to have this quality, as illustrated by the results. Finally, N-grams were severely affected by the top-$100$ token truncation. The best character n-gram model achieves macro/micro F-Measure scores of $0.507 / 0.603$ with NN classification and the best word n-gram model $0.493 / 0.627$ with KNN and NN classifiers. 

The results of the top individually performing features, in terms of micro / macro average F-Measure, are presented in the left half of table \ref{table:results}. \textbf{Bold} values represent column-wise maxima, while \underline{underlined} ones depict maxima in the left column category (e.g. feature type, in this case). ``NN\_ke'' and ``NN\_sk'' represent the keras and sklearn neural network implementations, repsectively. We can observe that the best performer is BOW with either LR or NNs, followed by word embeddings with NN classification. NGGs have a slightly worse performance, which can be attributed to the severely shorter (2D) feature vector it utilizes. On the other hand, BOW features are $1000$-dimensional vectors. Compared to NGGs, this corresponds to a $500$-fold dimension increase, with a $9.0 \%$ micro F-measure performance gain.

Subsequently, we test the question on whether the combination of features achieve a better performance than individual features.
The results are illustrated in the right half of Table \ref{table:results}. First, the \texttt{best} combination that involves NGG, BOW and GloVe features is, not surprisingly, the top performer, with LR and NN-sklearn obtaining the best performance. The \texttt{all} configuration follows with NB achieving macro/micro F-scores of $0.795$ and $0.792$ respectively. This shows that the additional features introduced significant amounts of noise, enough to reduce performance by canceling out any potential information the extra features might have provided. Finally, the \texttt{vector} combination achieves the worst performance: $0.787$ and $0.783$ in macro/micro F-measure. This is testament to the added value NGGs contribute to the feature pool, reinforced by the individual scores of the other vector-based approaches. 

\begin{table}
\caption{Average micro \& macro F-Measure for NGG, BOW and GloVe features (left) and the ``best'', ``vector'' and ``all'' feature combinations (right).}
\centering
\begin{tabular}
{|c|ccc||c|ccc|}
\hline
feature & classifier & macrof & microf & combo & classifiers & macrof  & microf  \\ \hline 
  \multirow{6}{*}{NGG}
         &  KNN  &  0.712  &  0.736                                              &  \multirow{6}{*}{best} &  KNN  &  0.810  &  0.820 \\                                                   
  &  LR  &  0.712  &  0.739                                                     &      &  LR  &  \underline{\textbf{0.819}}  &  \underline{\textbf{0.831}} \\
  &  NB  &  0.678  &  0.713                                                     &      &  NB  &  0.632  &  0.667 \\
  &  NN\_ke  &  \underline{0.718}  &  0.727                                  &      &  NN\_ke  &  0.807  &  0.819 \\
  &  NN\_sk  &  0.716  &  \underline{0.740}                                &      &  NN\_sk  &  \underline{\textbf{0.819}}  &  \underline{\textbf{0.831}} \\ 
  &  RF  &  0.699  &  0.726                                                     &      &  RF  &  0.734  &  0.759 \\ \hline
  \multirow{6}{*}{BOW}                                                           
         &  KNN  &  0.787  &  0.763                                             &   \multirow{6}{*}{all} &  KNN  &  0.497  &  0.569 \\                                                                   
  &  LR  &  {\textbf{0.808}}  &  \underline{0.776}                              &      &  LR  &  0.760  &  0.772 \\                                                      
  &  NB  &  0.629  &  0.665                                                     &      &  NB  &  \underline{0.795}  &  \underline{0.792} \\                                           
  &  NN\_ke  &  \underline{\textbf{0.808}}  &  \underline{0.776}             &      &  NN\_ke  &  0.537  &  0.629 \\                                                                
  &  NN\_sk  &  \underline{\textbf{0.808}}  &  \underline{0.776}           &      &  NN\_sk  &  0.664  &  0.678 \\                                                                
  &  RF  &  0.807  &  0.776                                                     &      &  RF  &  0.700  &  0.731 \\ \hline                                                            
  \multirow{6}{*}{glove}                                                                                                                            
  &  KNN  &  0.741  &  0.765                                                    &  \multirow{6}{*}{vector}  &  KNN  &  0.497  &  0.569 \\                                                                   
  &  LR  &  0.749  &  0.769                                                     &      &  LR  &  0.745  &  0.756 \\                                                                   
  &  NB  &  0.715  &  0.726                                                     &      &  NB  &  \underline{0.787}  &  \underline{0.783} \\                                           
  &  NN\_ke  &  0.774  &  0.788                                              &      &  NN\_ke  &  0.592  &  0.640 \\                                                                   
  &  NN\_sk  &  \underline{0.786}  &  \underline{\textbf{0.800}}           &      &  NN\_sk  &  0.669  &  0.675 \\                                                                
  &  RF  &  0.731  &  0.755                                                     &      &  RF  &  0.727  &  0.742 \\ \hline                                                            
\end{tabular}
\label{table:results}
\end{table}


Apart from experiments in the binary Hate Speech classification on the combined dataset, we have tested our classification models in multi-class classification, using the original RS and HSOL datasets. In RS, our best score was achieved with the \texttt{all} combination and the RF classifier with a micro F-Measure of $0.696$. For the HSOL dataset, we achieved a micro F-Measure of $0.855$, using the \texttt{best} feature combination and the LR classifier.

\subsection{Significance testing}

In table \ref{table:anova}  we present ANOVA results with respect to feature extractors and classifiers, under macro and micro F-measure scores. For both metrics, the selection of both features and classifiers is statistically significant with a confidence level greater than $99.9\%$.
We continue by performing a set of Tukey's Honest Significance Difference test experiments in table \ref{table:tukey}, depicting each statistically different group as a letter. In the upper part we present results  between feature combination groups (``a'' to ``d''), where the \texttt{best} combination is significantly different by the similar \texttt{all} and \texttt{vector} combinations by a large margin, as expected. The middle part compares individual features (groupped from ``a'' to ``g''), where GloVe, BoW and NGGs are assigned to neighbouring groups and arise the most significant features, with the other approaches having a large significance margin from them. Spelling and syntax features are grouped together, as well as the n-gram approaches.
Finally, the lower part of the table examines classifier groups (``a'' to ``c''). Here LR leads the ranking, followed by groups with the ANNs approaches, the NB and RF, and the KNN method.

\begin{table}
  \centering
\caption{ANOVA results with repect to feature and classifier selection, in terms of macro F-measure (left) and micro-Fmeasure (right).} 
\begin{tabular} {|l|c|c|} \hline


 parameter           &     Pr($>$F)  (macrof)  &      Pr($>$F)  (microf)              \\ \hline
features    &    \textbf{$<$ 2e-16} &     \textbf{$<$ 2e-16}              \\ 
classifiers &   \textbf{2.77e-05}   &     \textbf{8.65e-08}               \\ 
  \hline

\end{tabular}
\label{table:anova}
\end{table}

\subsection{Discussion}

The results and statistical tests on our work showcase the BOW, GloVe embeddings and the NGG model as the top performing feature-related configurations. BOW and GloVe score best in terms of micro and macro F-measure respectively, with NGG close behind, despite the extreme dimensionality reduction incurred by the model vector representation of graph similarities. The combination of the top performing features improves the results over individual ones, with 0.831 micro F-Measure when employed on an LR classifier or NN-sklearn.

\begin{table}
\caption{Tukey's HSD group test on micro F-Measure between feature combination groups (top), individual features (middle) and classifiers (bottom).}
  \centering
\begin{tabular} {|l|c|l|}
\hline
config  & micro F-measure & groups \\ \hline
best &  0.787 & a \\ 
all &  0.695 & \hspace{6pt}\hspace{6pt}cd \\ 
vector &  0.693 & \hspace{6pt}\hspace{6pt}\hspace{6pt}d \\ 
\hline
glove &  0.767 & a \\ 
BoW &  0.755 & ab \\ 
NGG & 0.730 & \hspace{6pt}bc \\ 
spelling & 0.617 & \hspace{6pt}\hspace{6pt}\hspace{6pt}\hspace{6pt}e \\ 
syntax & 0.613 & \hspace{6pt}\hspace{6pt}\hspace{6pt}\hspace{6pt}e \\ 
c-ngrams & 0.574 & \hspace{6pt}\hspace{6pt}\hspace{6pt}\hspace{6pt}\hspace{6pt}f \\ 
w-ngrams & 0.572 & \hspace{6pt}\hspace{6pt}\hspace{6pt}\hspace{6pt}\hspace{6pt}f \\ 
sentiment &  0.500 & \hspace{6pt}\hspace{6pt}\hspace{6pt}\hspace{6pt}\hspace{6pt}\hspace{6pt}g \\ 
\hline
LR & 0.689 & a \\
NN\_ke & 0.670 & ab \\
NN\_sk & 0.668 & ab \\
NB & 0.661 & \hspace{6pt}bc \\
RF & 0.655 & \hspace{6pt}bc \\
KNN & 0.639 & \hspace{6pt}\hspace{6pt}c \\ \hline
\end{tabular} 
\label{table:tukey}
\end{table}

Regarding classification methods, the LR and ANN classifiers perform best when used with our top performing features (separately or combined). Statistical tests show that in both micro and macro F-Measure terms, both representation and classification approaches have a significant role in the performance results. 

Finally, we understand from our study that the contribution of NGGs as a text representation is significant.
NGGs do not use domain-specific knowledge (unlike the BOW vectors which use HateBase keywords) nor require prior training on large document collections (unlike word embeddings, which need extensive unsupervised pre-training).
In addition, the vector dimension of the NGG-based approach is equal to the number of classes, as opposed to the $1000$ and $50$-dimensional BOW and embedding vectors, respectively. Despite this low dimensional representation, our empirical evaluation shows that NGGs have a significant contribution to detection performance.
Therefore NGGs can be seen as off-the-shelf rich features that encapsulate useful information in a low dimensional representation, which helps achieve significant performance either when used by itself or in feature combination approaches.

\section{Conclusion and Future Work}
\label{sec:Conclusion}
In this study, we investigated different text representation techniques and classification algorithms, performing a large number of experimental evaluations on the Hate Speech detection problem. We showed that n-gram graph-based features constitute deep/rich features, with significant contribution to the Hate Speech classification results.

In the future, we aim to better evaluate the contribution of word roles (e.g. POS tags) and combine them with improved preprocessing, to avoid possible noise in the related features. Concerning NGGs in Hate Speech detection, we want to apply the findings from the work of \cite{tsekouras2017graph} on NGG variations, to represent short texts with only the important n-grams of the text (e.g. through a TF-IDF filtering process and/or a named entity recognizer). The aim is to reduce the complexity and size of the NGGs, while retaining all the useful information. Another avenue of research, is the enrichment of deep features with statistical pre-trained models (such as Latent Dirichlet Allocation) or semantic information (e.g. from thesauri) to further improve performance.

\bibliographystyle{splncs04}
\bibliography{main}
\end{document}